\documentclass{article}
\usepackage[T1]{fontenc}
\usepackage{iclr2027_conference,times}
\iclrnamedcopy

\usepackage{amsmath,amsfonts,bm}

\def\eqref#1{equation~\ref{#1}}

\def\1{\bm{1}}

\DeclareMathAlphabet{\mathsfit}{\encodingdefault}{\sfdefault}{m}{sl}
\SetMathAlphabet{\mathsfit}{bold}{\encodingdefault}{\sfdefault}{bx}{n}

\usepackage{hyperref}
\hypersetup{hidelinks}
\usepackage{url}
\usepackage{graphicx}
\usepackage{float}
\usepackage{booktabs}
\usepackage{amsmath,amssymb}
\newcommand{\method}{PocketSplat}
\title{PocketSplat: Mobile Gaussian Reconstruction via World-Space Latent Allocation}
\author{%
Wenzhi Guo\thanks{Equal contribution. Emails: \href{mailto:wenzhi.guo@connect.polyu.hk}{\texttt{wenzhi.guo@connect.polyu.hk}} (Wenzhi Guo) and \href{mailto:cxdcsu@csu.edu.cn}{\texttt{cxdcsu@csu.edu.cn}} (Xianda Chen).}\And
Xianda Chen\footnotemark[1]\And
Dongxuan Chen\And
Guangchi Fang\And
Bing Wang\thanks{Corresponding author. Email: \href{mailto:bingwang@polyu.edu.hk}{\texttt{bingwang@polyu.edu.hk}}.}}
\begin{document}
\maketitle

\begin{figure}[H]
\centering
\includegraphics[width=\linewidth]{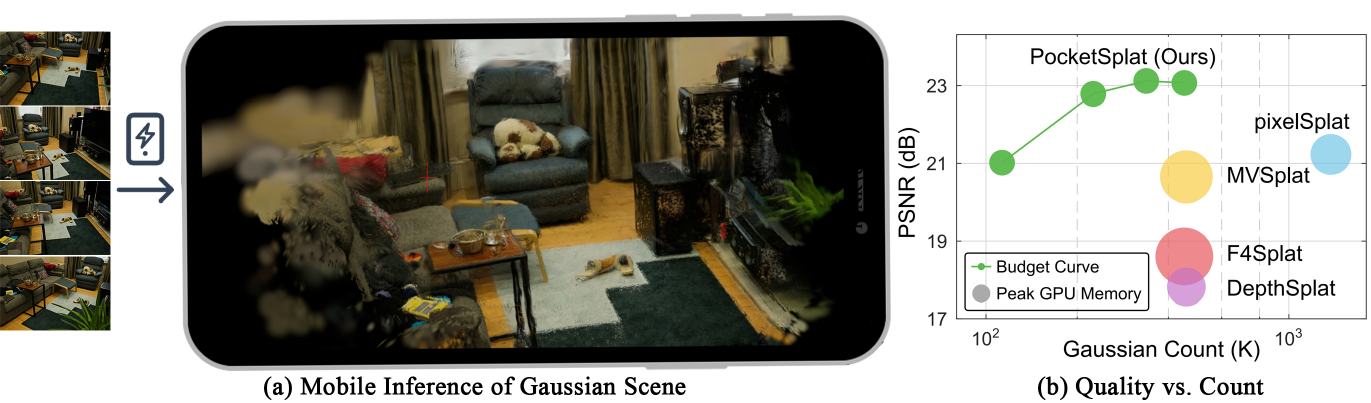}
\caption{\textbf{PocketSplat for mobile Gaussian reconstruction.} (a) Multi-view images are converted into a budgeted Gaussian representation and rendered on a mobile device. (b) The DL3DV quality--budget curve shows the reconstruction trade-off across representation budgets; bubble area denotes peak GPU memory.}
\label{fig:teaser}
\end{figure}

\begin{abstract}
Mobile Gaussian reconstruction must satisfy two requirements: the reconstruction model must execute within a device resource envelope, and the resulting Gaussian asset must expose a representation size suited to downstream mobile use. Existing feed-forward Gaussian reconstructors commonly decode dense, image-aligned candidates whose final cardinality is implicitly determined by the input resolution and number of views. We present PocketSplat, a feed-forward framework for budgeted mobile Gaussian asset construction. Given a prescribed output budget, PocketSplat organizes dense geometry-aware latent candidates in predicted world space, allocates exact integer capacity across local latent cells, and decodes complete Gaussian attributes only for retained candidates. Cell-conditioned latent fusion aggregates repeated multi-view evidence before decoding, while spatial responsibility decoding adapts Gaussian support after local sparsification. Experiments on DL3DV and out-of-distribution benchmarks establish a strong quality--budget trade-off against feed-forward Gaussian reconstruction baselines. On Mip-NeRF 360, PocketSplat executes directly on a target iPhone and constructs compact, higher-quality Gaussian assets substantially faster than a deployable streamed MVSplat variant; native MVSplat and DepthSplat exceed the device memory budget.
\end{abstract}

\section{Introduction}
Mobile 3D reconstruction would allow users to capture, revisit, and share everyday environments from a small set of images. Its practical value depends on two coupled requirements: the network must execute within device resources, and its Gaussian asset must remain usable after construction. Asset cardinality affects local storage, transmission, rasterization, and interactive processing, whereas the execution graph determines whether reconstruction can run on-device at all. Recent mobile systems address compact Gaussian rendering and representation processing \citep{du2026mobilegs}, while on-device optimization approaches expose coupled time and memory constraints \citep{guo2026pocketgs}. We therefore treat representation size as a first-class deployment variable. This paper studies an explicit \emph{output} budget: the prescribed number of Gaussian primitives made available to a downstream mobile application. It lets an application select a representation-size operating point for its storage, transfer, and rendering requirements, rather than accepting an asset size fixed implicitly by reconstruction inputs. This formulation makes representation capacity visible to the mobile application rather than an incidental by-product of input sampling.

Feed-forward Gaussian reconstruction is a promising route to this setting. 3D Gaussian Splatting represents a scene with anisotropic, differentiably rasterized primitives \citep{kerbl2023gaussian}; generalizable predictors amortize reconstruction across scenes rather than optimizing a representation from scratch for every capture. pixelSplat infers Gaussians from image pairs \citep{charatan2024pixelsplat}, MVSplat incorporates multi-view stereo \citep{chen2024mvsplat}, and DepthSplat couples Gaussian prediction with learned depth \citep{xu2025depthsplat}. However, dense image-aligned prediction commonly ties the final Gaussian population to image resolution and input views. Repeated observations of one local region can thus create many latent candidates before a representation decision is made. Dense multi-view intermediates can also be difficult to execute directly within a mobile memory envelope. Applying a budget after independently decoding every candidate leaves capacity allocation and asset size as secondary decisions. It also prevents an application from selecting its desired quality--asset-size operating point before scene construction begins.

Controllable-count and adaptive-density methods have begun to address representation size. SparseSplat predicts a sparse set of pixel-unaligned primitives but does not enforce an exact output budget or adapt support to local compression \citep{zhang2026sparsesplat}; EcoSplat adapts density in image space without associating candidates across views \citep{park2025ecosplat}; F$^4$Splat controls count through threshold-based pruning after decoding, without fusing grouped observations or adapting primitive support to local compression \citep{kim2026f4splat}. Under a prescribed budget, three decisions remain coupled. First, image-space candidates must be associated in a common world-space structure despite viewpoint changes and imperfect geometry. Second, a retained candidate should incorporate information from repeated observations that will not become output primitives. Third, reducing local density changes how much of the scene a surviving primitive represents. Pixel ranking and post-hoc pruning do not coordinate these decisions. We instead formulate budgeted reconstruction as \emph{world-space latent allocation}: distribute a prescribed population across predicted world-space latent cells, aggregate cell observations, and adapt retained support to local compression.

We present \method, a budget-constrained feed-forward framework for mobile Gaussian asset construction. A frozen multi-view geometry backbone supplies geometric priors and intermediate features, from which a trainable head produces dense geometry-aware latent candidates. We lift candidates to predicted world space and group them into latent cells, so repeated observations participate in a common allocation decision. The allocator assigns an exact feasible budget using local image detail, view-mean weighting that discounts repeated evidence, a support floor for zero-detail cells, and capacity completion. For each retained candidate, hidden fusion aggregates weighted cell information before complete Gaussian attributes are decoded. Spatial responsibility decoding (SRD) then modifies Gaussian scale through an analytic local-compression prior and a learned residual. Every operating point returns exactly its requested number of Gaussians while allocating capacity across predicted scene structure. The resulting asset cardinality is therefore independent of the dense candidate field used for geometric reasoning. It decouples dense geometric inference from the compact representation delivered to the mobile application after construction.

Our contributions are:
\begin{itemize}
\item The first feed-forward 3D Gaussian reconstruction framework for mobile asset construction, with an explicit output budget that exposes controllable representation cardinality for downstream mobile use.
\item A world-space exact-budget allocator that combines local detail, cross-view redundancy weighting, zero-detail support, and capacity completion before attribute decoding.
\item Cross-view hidden fusion and scale-only spatial responsibility decoding that adapt retained primitives to local sparsification, with in-domain, out-of-domain, and on-device evaluation.
\end{itemize}

\section{Related Work}
\paragraph{Gaussian representations and feed-forward reconstruction.}
Novel-view synthesis progressed from neural radiance fields and efficient grid or hash encodings \citep{mildenhall2020nerf,muller2022instant} to 3D Gaussian Splatting, which enables real-time rasterization after scene-specific optimization \citep{kerbl2023gaussian}. Subsequent work improves Gaussian fidelity and structure through anti-aliasing, anchors, and surface-oriented representations \citep{yu2023mipsplatting,lu2023scaffoldgs,huang2024twodgs,guedon2024sugar,chen2024pgsr,yu2024gof}, while pruning, quantization, coding, and constrained optimization compact optimized scenes \citep{fan2024lightgaussian,niedermayr2024compressed,navaneet2023compgs,chen2024hac,fang2024minisplatting,mallick2024taming}. Generalizable reconstruction instead predicts renderable Gaussians without per-scene optimization, spanning monocular, image-pair, multi-view, variational, transformer, depth-guided, pose-free, and unconstrained settings \citep{szymanowicz2024splatter,szymanowicz2024flash3d,charatan2024pixelsplat,chen2024mvsplat,wewer2024latentsplat,zhang2024gslrm,xu2025depthsplat,smart2024splatt3r,ye2024noposplat,zhang2025flare,jiang2025anysplat}. Sparse and controllable-count approaches further adapt primitive placement, alignment, anchors, or density \citep{zhang2026sparsesplat,moreau2026offgrid,zhang2026anchorsplat,park2025ecosplat,kim2026f4splat}. PocketSplat complements these approaches by allocating an exact output population over grouped world-space latent candidates before attribute decoding.

\paragraph{Geometry priors and mobile deployment.}
Geometry foundation models provide depth, point-map, matching, and camera priors \citep{wang2024dust3r,leroy2024mast3r,zhang2024monst3r,yang2025fast3r,yin2023metric3d,piccinelli2024unidepth,yang2024depthanythingv2,wang2025vggt,depthanything3}. Efficiency work makes neural representations practical through baked grids, memory-efficient rendering, mobile rasterization, optimized Gaussian rasterizers, pruning, and localization \citep{hedman2021snerg,reiser2023merf,chen2023mobilenerf,feng2024flashgs,hanson2024speedysplat,hanson2025pup3dgs}; Mobile-GS and PocketGS address mobile Gaussian inference and on-device optimization \citep{du2026mobilegs,guo2026pocketgs}. PocketSplat uses frozen geometry priors to construct a budgeted Gaussian asset, combining direct mobile execution with explicit control of downstream representation cardinality across device-specific storage, transfer, rendering, and interactive processing requirements.

\section{Method}
\label{sec:method}
\begin{figure*}[t]
\centering
\includegraphics[width=\textwidth]{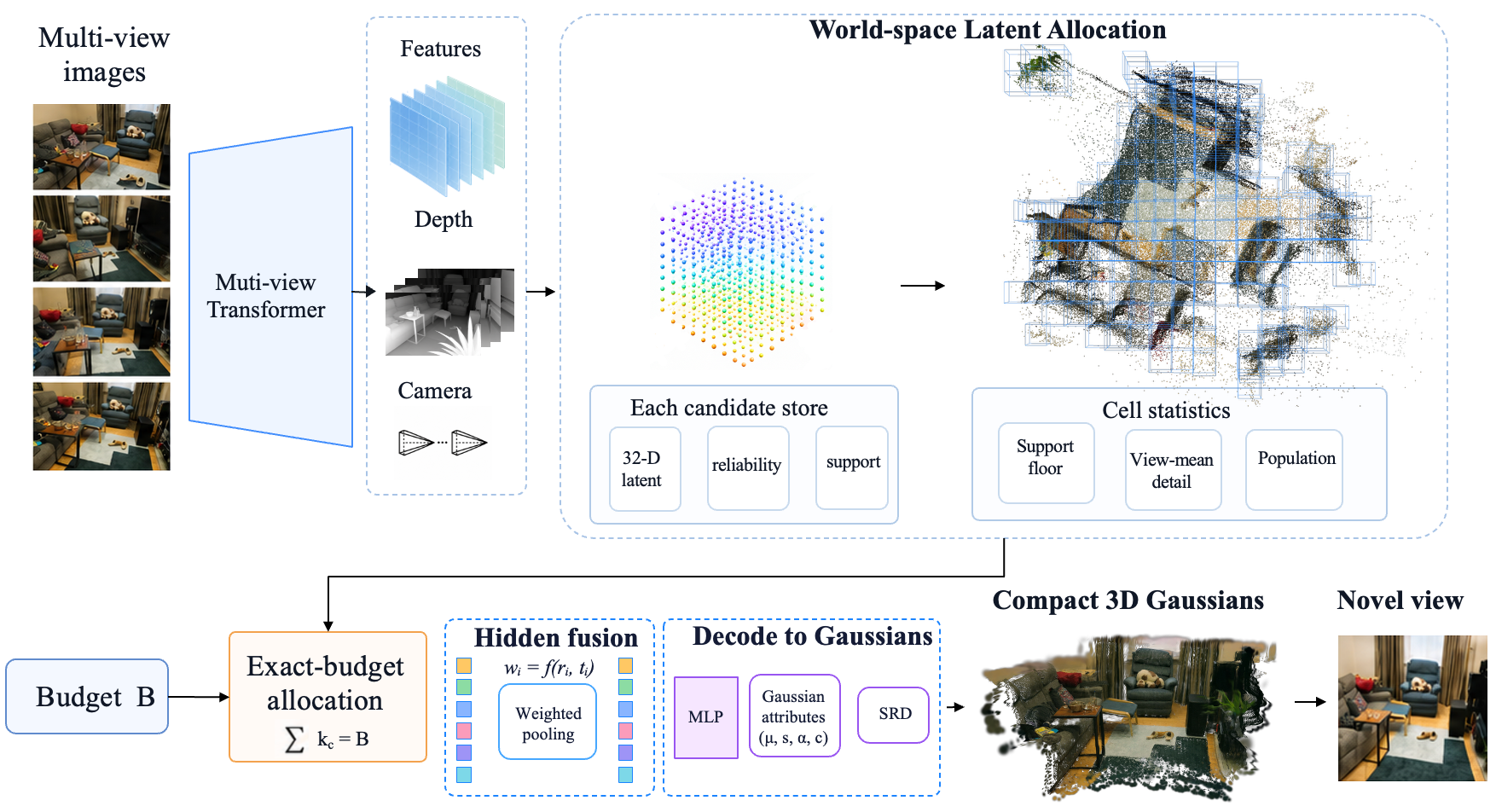}
\caption{\textbf{PocketSplat overview.} Multi-view images enter a frozen multi-view transformer, yielding features, depth, and cameras. Each dense candidate stores a latent, reliability, and support. World-space anchors form cells with population, support-floor, and view-mean-detail statistics. Given $B$, exact-budget allocation returns quotas satisfying $\sum_c k_c=B$. Weighted hidden fusion pools each cell, after which retained candidates are decoded and adapted by SRD to form compact Gaussians for novel-view rendering.}
\label{fig:pipeline}
\end{figure*}

\method{} is organized around an image-conditioned world-space latent field and a budget-conditioned Gaussian decoder. Given $V$ multi-view images with intrinsics and camera-to-world poses, denoted by $\mathcal X_c$, and a feasible Gaussian budget $B$, it constructs
\begin{equation}
\mathcal G_B=f_\theta(\mathcal X_c,B)
 =\{\mu_i,\Sigma_i,c_i,o_i\}_{i\in\mathcal S_B},\qquad |\mathcal S_B|=B.
\label{eq:budgeted_asset}
\end{equation}
Here $\mu_i\in\mathbb R^3$, $\Sigma_i\in\mathbb R^{3\times3}$, $c_i$, and $o_i$ denote a world-space mean, covariance, degree-2 RGB spherical-harmonic coefficients, and opacity logit. The latent field consolidates dense multi-view geometric evidence before materialization; the decoder converts it into a mobile Gaussian asset whose cardinality is specified by $B$. Figure~\ref{fig:pipeline} summarizes this representation-to-asset construction.

\subsection{World-space latent field}
A frozen multi-view transformer, camera modules, and a depth head produce four feature levels together with predicted depth, intrinsics, and world-to-camera transforms. A trainable GS head converts these outputs to a 32-channel hidden map and predicts geometry-related reliability and cross-view support. We instantiate the dense image-conditioned field as
\begin{equation}
\mathcal Z=\{(x_i,h_i,r_i,t_i)\}_{i=1}^{N},
\end{equation}
where $x_i$ is a predicted world-space anchor and $(h_i,r_i,t_i)$ are the candidate feature, reliability, and support. This is an allocation-time representation rather than a renderable Gaussian set: complete Gaussian attributes are produced only for the active support selected by the decoder. At runtime, $V=4$, $H=252$, and $W=448$, giving $N=VHW=451{,}584$ dense candidates. The same field supports every requested representation budget.

\paragraph{Latent structure.}
For a pixel $p=(u,v,1)^\top$ in multi-view image $j$, let $\hat D_j$, $\hat K_j$, and $\hat E_j$ denote predicted depth, intrinsics, and world-to-camera pose. We first unproject the pixel in camera coordinates and transform it to world coordinates:
\begin{equation}
 x_i^c=\hat D_j(u,v)\hat K_j^{-1}p,\qquad
 \begin{bmatrix}x_i\\1\end{bmatrix}
 =\hat E_j^{-1}\begin{bmatrix}x_i^c\\1\end{bmatrix}.
\label{eq:unproject}
\end{equation}
The median finite positive multi-view depth $d_{\rm scene}$ defines a scene-relative cell width $v_s=0.02d_{\rm scene}$. Candidates are assigned by $b_i=\lfloor x_i/v_s\rfloor$, yielding world-space cells $\mathcal C_c=\{i:b_i=c\}$ with populations $n_c=|\mathcal C_c|$. These cells provide the common unit for allocation and feature aggregation, while all decoded means $\mu_i$ remain continuous. By grouping in world space, observations that project to distant image coordinates but correspond to a common local region can participate in the same representation decision.

\paragraph{Cell statistics.}
The allocation policy gives capacity to cells according to local visual detail and cross-view evidence. Candidate detail $e_i$ is the entropy of a $7\times7$ multi-view RGB neighborhood. Image preprocessing follows the fixed convention in Appendix~\ref{app:allocation_impl}. Given local histogram probability $p_{i,b}$, detail is
\begin{equation}
e_i=-\sum_b p_{i,b}\log p_{i,b},
\end{equation}
with zero terms omitted. Let $\bar e_c$ be the mean detail in cell $c$. We define a zero-detail support floor, positive-detail capacity, detail mass, view count, and view-mean weight as
\begin{align}
 l_c&=\mathbf 1[n_c>0\ \land\ \bar e_c=0],& m_c&=n_c-l_c,\\
 W_{\rm sum}(c)&=m_c\bar e_c,&
 V_c&=\left|\left\{j:\sum_{i\in\mathcal C_c\cap\mathrm{view}(j)}e_i>0\right\}\right|,\\
 W_{\rm vm}(c)&=\begin{cases}W_{\rm sum}(c)/V_c,&W_{\rm sum}(c)>0,\\0,&\text{otherwise}.\end{cases}
\label{eq:detail_weights}
\end{align}
The support floor retains one representative for an occupied zero-detail cell. The view-mean term discounts detail evidence replicated in multiple multi-view images, so capacity is assigned according to scene content rather than repeated image-plane evidence alone.

\subsection{Budget-conditioned Gaussian asset decoding}
\paragraph{Exact-budget support.}
Let $L=\sum_c l_c$ and $R=B-L$. For feasible support constraints ($L\le B\le N$), residual quotas for positive-detail cells are
\begin{equation}
 q_{{\rm res},c}=\min(m_c,\lambda W_{\rm vm}(c)),\qquad q_c=l_c+q_{{\rm res},c},
\label{eq:waterfill}
\end{equation}
where active-set water filling sets $\lambda$ such that residual quotas sum to $R$. When positive-detail cells reach capacity, V2 capacity completion distributes remaining slots over the residual capacity of zero-detail cells. Corrected stochastic rounding then returns integer cell quotas
\begin{equation}
 l_c\le k_c\le n_c,\qquad \sum_c k_c=B.
\label{eq:exact_budget}
\end{equation}
Thus, every feasible operating point produces the exact requested asset cardinality without leaving budget slots unused.

Within each cell, reproducible keys select $k_c$ distinct candidates. At the dense operating point $B=N$, the allocator returns $\mathcal S_N=\operatorname{arange}(N)$ and $k_c=n_c$ for every cell; Appendix~\ref{app:allocation_impl} gives the key construction.

\paragraph{Cell-conditioned Gaussian decoding and SRD.}
After allocation, we aggregate evidence within each cell before decoding its retained candidates. Normalized reliability $r_i$ and support $t_i\in[0,1]$ define pooling weights
\begin{equation}
w_i=(0.25+0.75r_i)(0.5+0.5t_i),
\end{equation}
The pooled latent and selected feature update are
\begin{equation}
 \bar h_c=\frac{\sum_{i\in\mathcal C_c}w_i h_i}{\sum_{i\in\mathcal C_c}w_i},\qquad
 h'_i=h_i+g\odot(\bar h_c-h_i),\quad g=0.5\tanh(\theta_g).
\label{eq:fusion}
\end{equation}
The learned channel-wise gate $g$ transfers cell-level evidence to selected candidates. Complete Gaussian attributes are projected from $h'_i$, so unselected candidates remain latent and do not enter the output asset.

Reducing a cell from $n_c$ candidates to $k_c$ changes the local support assigned to every retained primitive. For $k_c>0$, we define the local compression ratio and the SRD parameters as
\begin{equation}
 \rho_c=\log\frac{n_c}{k_c},\qquad
 a_i=1+\tanh(u_i),\qquad \delta_i=0.25\tanh(v_i),
\end{equation}
where a learned SRD head predicts $u_i$ and $v_i$. After decoding a physical base scale, SRD applies
\begin{equation}
 \gamma_i=\exp\left(\frac{a_i}{2}\rho_c+\delta_i\right),\qquad
 s_i'=\gamma_i s_i,\qquad \Sigma_i'=\gamma_i^2\Sigma_i.
\label{eq:srd}
\end{equation}
The factor $\sqrt{n_c/k_c}$ embedded in Eq.~\ref{eq:srd} allocates a larger local area responsibility to each retained Gaussian as the cell becomes sparse. The learned gate and residual adapt this geometric prior to the local scene. SRD modifies covariance after selection and leaves centers, rotations, opacity, spherical-harmonic appearance, image-plane offsets, and depth offsets unchanged.

\subsection{Training}
\label{sec:training}
We train on DL3DV chunks for 50,000 steps with batch size two, AdamW ($2\times10^{-4}$ learning rate and 0.01 weight decay), gradient clipping at 0.5, and VGG LPIPS. The multi-view geometry backbone remains frozen; the GS head, allocation-related modules, fusion module, and SRD head are optimized with RGB reconstruction, perceptual, and budget-selection supervision:
\begin{equation}
\mathcal L=\mathcal L_{\rm MSE}+\lambda_{\rm LPIPS}\mathcal L_{\rm LPIPS}+\lambda_{\rm sel}\mathcal L_{\rm selection}.
\end{equation}
Training uses the Full budget for the first 10,000 steps and then cycles through descending budgets with 100-step holds. The geometry grid is $252\times448$ after processing the configured $256\times448$ bucket. The iOS fixed-shape decoder is described separately in Appendix~\ref{app:mobile_deployment}.

\section{Experiments}
\label{sec:evidence}
We evaluate the evidence chain behind budgeted mobile Gaussian asset construction. We first measure reconstruction quality at exact Gaussian budgets, then test transfer beyond the training distribution, profile construction on GPU, and evaluate direct deployment on a target iPhone. Ablations isolate the contribution of world-space allocation, spatial responsibility decoding, and allocation before complete attribute projection. This organization connects the evaluation to controllable assets, reconstruction quality, and feasible mobile execution. The resulting evaluation directly links each empirical result to a component of the proposed budgeted construction pipeline.

\subsection{Experimental setup}
\paragraph{Datasets and protocol.}
The in-domain benchmark uses DL3DV-Benchmark, the official test set of the DL3DV dataset. Each scene uses four multi-view images and 50 evaluation frames at $252\times448$. We evaluate frozen checkpoints out of distribution on Tanks \& Temples and MegaDepth. Section~\ref{sec:mobile_eval} reports a separate on-device evaluation on Mip-NeRF 360. Together, these settings test in-distribution reconstruction, transfer to unfamiliar scenes, and deployed mobile behavior. They also expose the same model family to both standard benchmark scenes and real mobile capture sequences.

\paragraph{Baselines, budgets, and metrics.}
We compare pixelSplat, MVSplat, DepthSplat, TranSplat, F$^4$Splat, and \method. The first four use their native Gaussian counts and F$^4$Splat uses its native threshold-based count mechanism. One \method{} checkpoint is evaluated at $B\in\{113{,}000,226{,}000,338{,}000,451{,}584\}$. We report PSNR after clipping RGB to $[0,1]$, SSIM with an 11-pixel Gaussian window, and VGG LPIPS. Metrics are averaged within each scene and then over scene means. SRD confidence intervals use paired scene bootstrapping with 10,000 resamples. Training settings appear in Section~\ref{sec:training}. All evaluations use these definitions to report reconstruction quality across the stated operating configurations. This protocol makes the reported reconstruction measures directly comparable across the budget settings of the proposed method.

\subsection{Reconstruction Quality Across Datasets}
\paragraph{In-domain quality and exact budget control.}
Table~\ref{tab:main_all} establishes the quality--budget operating curve on DL3DV. At 226K, \method{} uses 16.4\% of pixelSplat's reported Gaussian count while reaching 22.790 dB, a 1.559 dB PSNR improvement. Increasing the budget from 226K to 338K adds 0.332 dB PSNR and reduces LPIPS from 0.2142 to 0.1887. The 338K configuration uses 25.2\% fewer Gaussians than Full while changing PSNR by only $+0.0526$ dB and SSIM by $-0.0034$; Full provides the lowest LPIPS at 0.1776. Even Light at 113K reaches 21.018 dB, exceeding pixelSplat while using 8.2\% of its reported count. \mbox{This establishes high-quality operating points for compact mobile Gaussian assets.}

\begin{figure}[t]
\centering
\includegraphics[width=\linewidth]{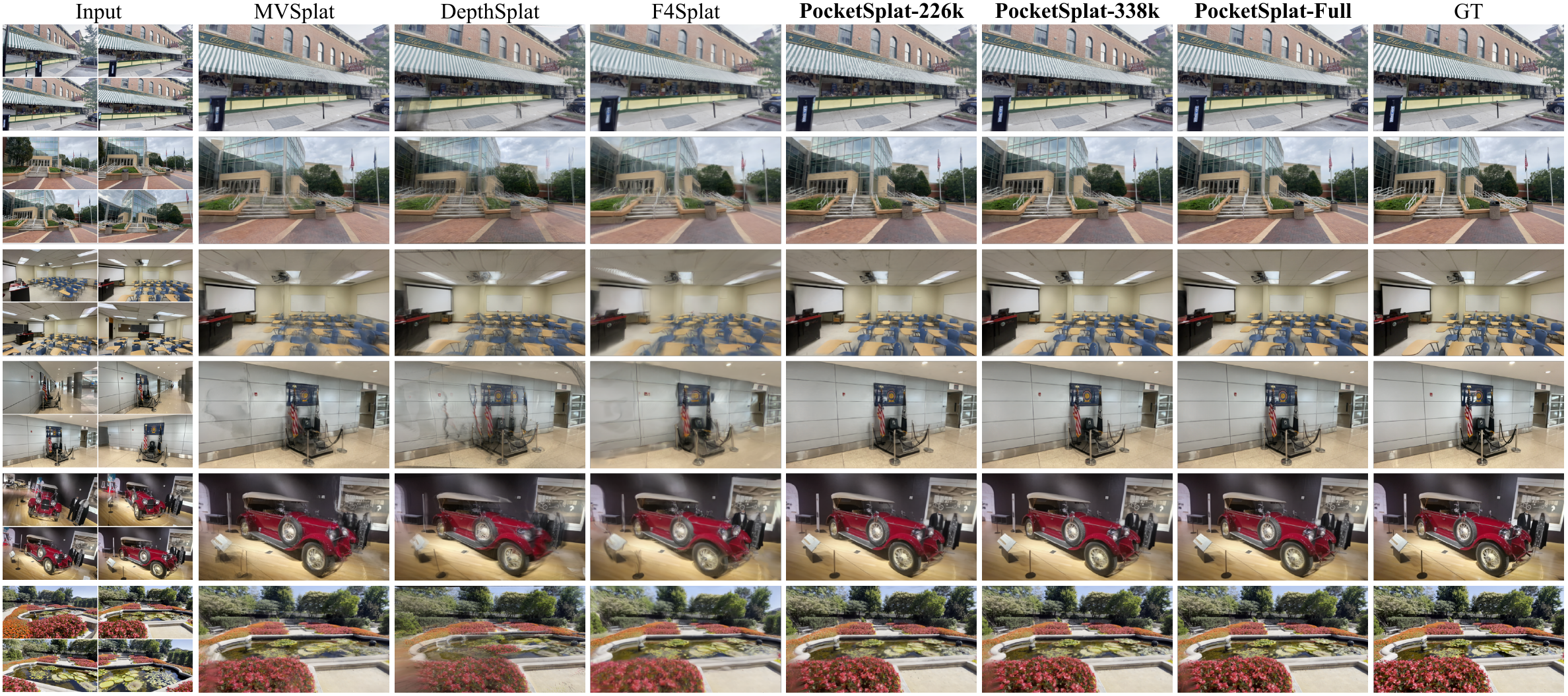}
\caption{Qualitative DL3DV comparison with six examples. Columns show context inputs, MVSplat, DepthSplat, F$^4$Splat, PocketSplat at 226K, 338K, and Full, and ground truth.}
\label{fig:qual_dl3dv}
\end{figure}

The comparison also separates exact allocation from threshold-driven count control. At similar 113K, 226K, and 338K populations, PocketSplat exceeds F$^4$Splat by 2.477, 4.196, and 4.518 dB, respectively. At 338K, it improves over pixelSplat, MVSplat, DepthSplat, and TranSplat by 1.891, 2.467, 5.305, and 2.454 dB. Figure~\ref{fig:qual_dl3dv} visualizes this progression: the 226K and 338K assets preserve scene layout and appearance while exposing materially smaller representation sizes. Together, the table and figure show that the requested budget selects a usable quality--asset-size operating point rather than merely discarding a fixed dense output.

\begin{table*}[t]
\caption{Reconstruction quality across in-domain and out-of-distribution datasets. Higher PSNR/SSIM and lower LPIPS are preferable.}
\label{tab:main_all}
\centering\scriptsize\setlength{\tabcolsep}{3.0pt}
\begin{tabular}{lr rrr rrr rrr}
\toprule
& & \multicolumn{3}{c}{DL3DV} & \multicolumn{3}{c}{Tanks \& Temples} & \multicolumn{3}{c}{MegaDepth}\\
\cmidrule(lr){3-5}\cmidrule(lr){6-8}\cmidrule(l){9-11}
Method & Gaussians & PSNR$\uparrow$ & SSIM$\uparrow$ & LPIPS$\downarrow$ & PSNR$\uparrow$ & SSIM$\uparrow$ & LPIPS$\downarrow$ & PSNR$\uparrow$ & SSIM$\uparrow$ & LPIPS$\downarrow$\\
\midrule
pixelSplat & 1,376,256 & 21.231 & 0.7119 & 0.3463 & 15.839 & 0.3968 & 0.5665 & 16.213 & 0.5036 & 0.5079\\
MVSplat & 458,752 & 20.655 & 0.6646 & 0.2694 & 14.783 & 0.3471 & 0.4971 & 14.779 & 0.4218 & 0.4671\\
DepthSplat & 458,752 & 17.817 & 0.5370 & 0.3508 & 13.722 & 0.3224 & 0.5581 & 14.450 & 0.4079 & 0.4869\\
TranSplat & 458,752 & 20.668 & 0.6856 & 0.2630 & 14.692 & 0.3423 & 0.4987 & 14.880 & 0.4441 & 0.4525\\
\midrule
F$^4$Splat (113K) & 111,234 & 18.541 & 0.5230 & 0.3927 & 12.910 & 0.3181 & 0.6146 & 15.359 & 0.4230 & 0.5247\\
F$^4$Splat (226K) & 222,468 & 18.594 & 0.5275 & 0.3741 & 12.911 & 0.3177 & 0.6031 & 15.351 & 0.4226 & 0.5092\\
F$^4$Splat (338K) & 332,718 & 18.604 & 0.5292 & 0.3645 & 12.911 & 0.3183 & 0.5962 & 15.365 & 0.4227 & 0.5013\\
F$^4$Splat (Full) & 451,584 & 18.604 & 0.5288 & 0.3619 & 12.876 & 0.3197 & 0.5913 & 15.337 & 0.4211 & 0.4985\\
\midrule
\method (113K) & 113,000 & 21.018 & 0.6690 & 0.2952 & 17.371 & 0.4621 & 0.4285 & 16.550 & 0.4952 & 0.4382\\
\method (226K) & 226,000 & 22.790 & 0.7399 & 0.2142 & 17.965 & 0.4947 & 0.3711 & 17.178 & 0.5335 & 0.3806\\
\method (338K) & 338,000 & 23.122 & 0.7561 & 0.1887 & 17.956 & 0.4982 & 0.3476 & 17.144 & 0.5396 & 0.3562\\
\method (Full) & 451,584 & 23.070 & 0.7595 & 0.1776 & 17.712 & 0.4925 & 0.3335 & 16.626 & 0.5274 & 0.3450\\
\bottomrule
\end{tabular}
\end{table*}

\begin{figure}[t]
\centering
\includegraphics[width=\linewidth]{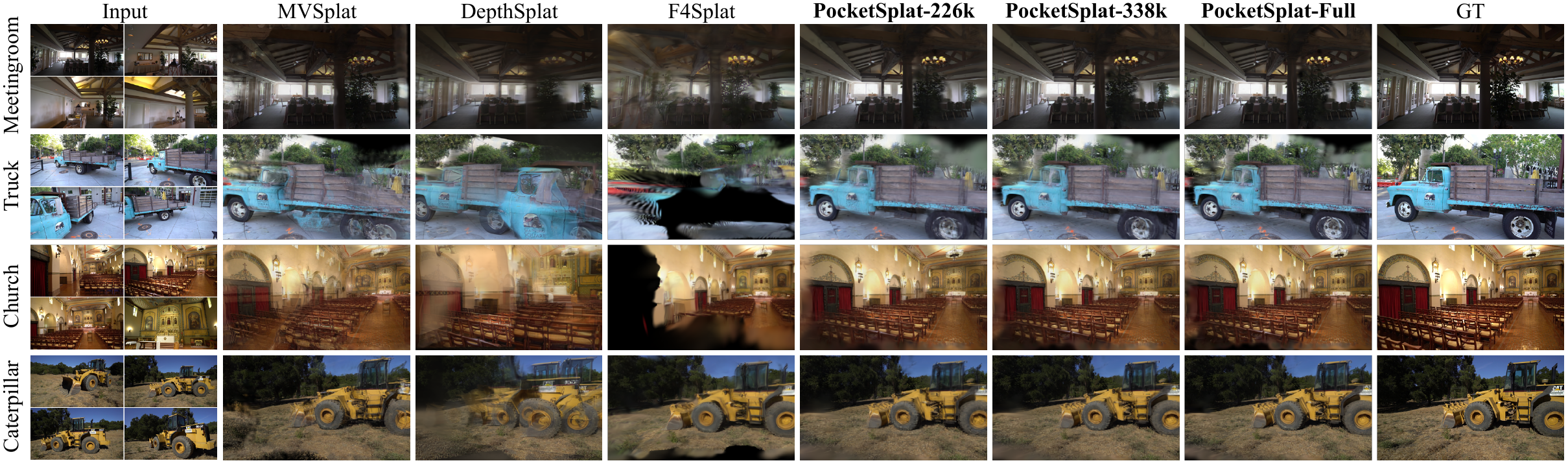}
\caption{Tanks \& Temples qualitative comparison on Meetingroom, Truck, Church, and Caterpillar. Columns show context inputs, MVSplat, DepthSplat, PocketSplat at 226K, 338K, and Full, and ground truth.}
\label{fig:qual_tnt}
\end{figure}

\paragraph{Cross-dataset generalization.}
Frozen-checkpoint evaluation on Tanks \& Temples and MegaDepth retains the budget-controlled behavior beyond DL3DV. PocketSplat PSNR peaks at 226K, reaching 17.965 dB on Tanks \& Temples and 17.178 dB on MegaDepth; Full gives the lowest LPIPS, 0.3335 and 0.3450. The 226K configuration improves on the strongest listed baseline PSNR by 2.126 dB on Tanks \& Temples and 0.965 dB on MegaDepth. At 338K, PocketSplat maintains comparable PSNR while reducing LPIPS to 0.3476 and 0.3562. Figure~\ref{fig:qual_tnt} shows the resulting reconstructions on Meetingroom, Truck, Church, and Caterpillar, and Table~\ref{tab:main_all} gives the complete measurements. These results show that the same allocation rule transfers a family of representation budgets across datasets. The budget-control behavior thus extends beyond the scenes used to train the reconstruction model. This consistency supports the use of one checkpoint across the reported representation-size operating points. It shows that the allocation mechanism remains useful when scene content and capture conditions differ from the training distribution.

\subsection{Mobile Evaluation}
\label{sec:mobile_eval}
\paragraph{Protocol and deployment path.}
We deploy on Mip-NeRF 360 \citep{barron2022mipnerf360}. Frames 0, 2, 4, and 6 provide four context inputs for each sequence; evaluation uses the first 20 remaining frames. All methods receive $256\times448$ RGB inputs with shared resize, crop, and camera preprocessing. The four PocketSplat user settings---Light, Balanced, Quality, and Dense---correspond to increasing output budgets. We use a shared FP32 hybrid execution path with Core ML visual backbones and ONNX Runtime for remaining computation. This protocol fixes the available multi-view evidence and output resolution for every reported mobile result.

PocketSplat executes directly through its standard Core ML and ONNX Runtime path. Native MVSplat and DepthSplat exhaust memory on the target iPhone because their multi-view cost volumes materialize depth candidates. We therefore report MVSplat-Stream: an engineering-modified MVSplat path that streams eight depth candidates at a time, reduces each chunk to channel correlations, writes it to the final cost-volume buffer, and releases its transient storage before processing the next chunk. Its Core ML backbone runs on CPU and remaining stages use ONNX Runtime. MVSplat-Stream completes deployment, whereas DepthSplat remains OOM after streamed cost-volume and staged execution. Appendix~\ref{app:mobile_deployment} and Table~\ref{tab:mobile_deployment} detail these direct-deployment outcomes.

\begin{table*}[t]
\centering\small\setlength{\tabcolsep}{3.0pt}
\caption{On-device Mip-NeRF 360 evaluation. Times are seconds, peak footprint is reported in GB, and higher PSNR/SSIM and lower LPIPS are preferable.}
\label{tab:mobile_results}
\begin{tabular}{llrrrrrrr}
\toprule
Method / setting & Status & Gaussians & Reconstruction & Inference & Peak & PSNR & SSIM & LPIPS\\
\midrule
MVSplat & OOM & 458,752 & -- & -- & -- & -- & -- & --\\
DepthSplat & OOM & 458,752 & -- & -- & -- & -- & -- & --\\
MVSplat-Stream & Deployable & 458,752 & 42.00 & 40.16 & 1.99 & 17.875 & 0.5793 & 0.3165\\
\midrule
PocketSplat Light & Deployable & 112,896 & 10.67 & 10.00 & 2.68 & 20.018 & 0.6690 & 0.2952\\
PocketSplat Balanced & Deployable & 225,792 & 10.20 & 9.24 & 2.89 & 21.790 & 0.6343 & 0.2142\\
PocketSplat Quality & Deployable & 338,000 & 10.95 & 9.59 & 2.64 & 22.132 & 0.6503 & 0.1923\\
PocketSplat Dense & Deployable & 451,584 & 10.50 & 9.27 & 2.66 & 22.034 & 0.6491 & 0.1897\\
\bottomrule
\end{tabular}
\end{table*}

\paragraph{Quality, time, and asset size.}
Table~\ref{tab:mobile_results} jointly measures direct deployability, representation size, construction time, inference time, peak footprint, and reconstruction quality. PocketSplat runs at every budget, constructing assets in 10.20--10.95 s and completing inference in 9.24--10.00 s. Balanced uses half the Dense primitive population, gives the lowest reconstruction and inference times, and reaches 21.790 dB. Quality reaches the strongest PSNR and SSIM, 22.132 dB and 0.6503, while Dense gives the lowest LPIPS, 0.1897. These settings let a mobile application select between compactness and reconstruction fidelity without changing the deployed model.

\begin{figure}[t]
\centering
\includegraphics[width=\linewidth]{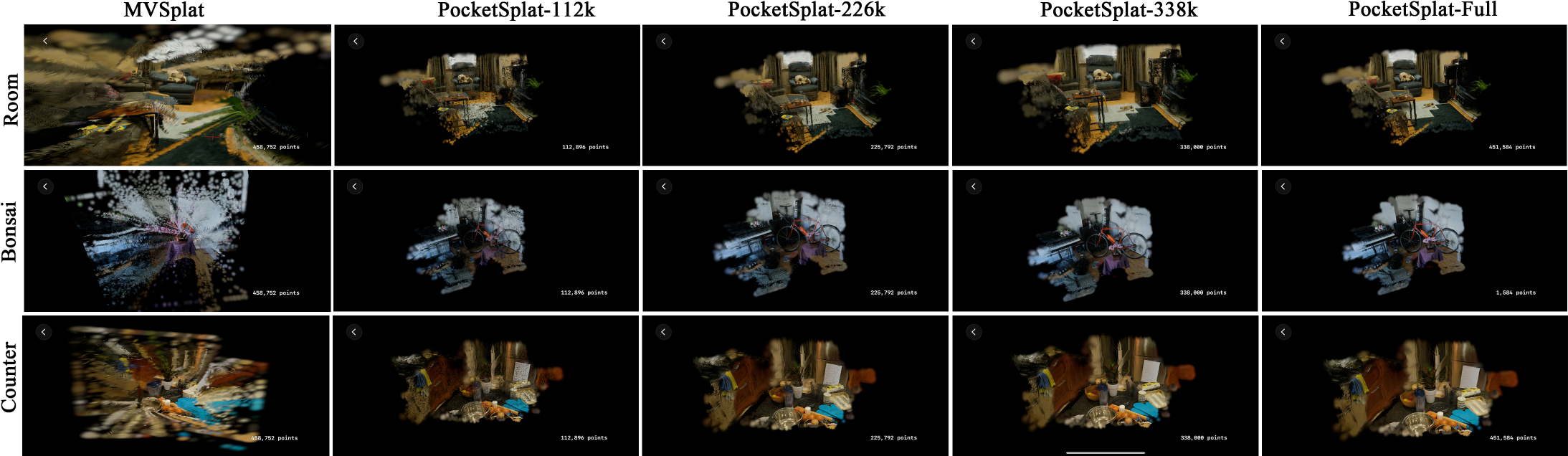}
\caption{Mobile visualizations on Mip-NeRF 360: Room, Bonsai, and Counter (rows), comparing MVSplat-Stream and PocketSplat at Light, Balanced, Quality, and Dense budgets (columns). PocketSplat budgets are 112,896, 225,792, 338,000, and 451,584, as reported in Table~\ref{tab:mobile_results}.}
\label{fig:mobile_results}
\end{figure}

MVSplat-Stream uses 458,752 Gaussians, takes 42.00 s for reconstruction and 40.16 s for inference, and reaches 17.875 dB. In contrast, PocketSplat Light uses 75.4\% fewer Gaussians while improving PSNR by 2.143 dB, SSIM by 0.0897, and LPIPS by 0.0213. PocketSplat Quality improves PSNR by 4.257 dB and SSIM by 0.0710 over MVSplat-Stream, with 338,000 output primitives. PocketSplat Dense reduces construction and inference time by 75.0\% and 76.9\%, respectively, relative to MVSplat-Stream. Its peak footprint ranges from 2.64 to 2.89 GB across user budgets, while the unmodified MVSplat and DepthSplat paths do not complete within the target-device memory envelope. Figure~\ref{fig:mobile_results} shows the corresponding progression on Room, Bonsai, and Counter.

\subsection{Ablation studies}
We isolate the components that turn the dense candidate field into a budgeted Gaussian asset on DL3DV. The following studies identify the contributions of allocation, support adaptation, and decode ordering. Each ablation holds the requested budget fixed while changing the corresponding construction mechanism or adaptation component.

\paragraph{World-space allocation.}
Uniform random candidate selection preserves the requested output count but ignores scene structure. World-space allocation improves PSNR by 3.0049 dB at 113K, 3.6915 dB at 226K, and 2.1916 dB at 338K (Table~\ref{tab:ablation}). The corresponding Random-to-PocketSplat changes are 18.0134 to 21.0183 dB, 19.0984 to 22.7899 dB, and 20.9305 to 23.1221 dB. These results isolate the benefit of assigning fixed capacity across world-space latent cells instead of selecting a count-matched random subset. The gain at every sparse operating point shows that count alone does not determine representation quality. Allocating the same primitive population according to predicted world-space structure is therefore essential at tight budgets.

\begin{table}[t]
\caption{Quality ablations on DL3DV (PSNR, dB).}
\label{tab:ablation}
\centering\small\setlength{\tabcolsep}{4pt}
\begin{tabular}{rrrrr}
\toprule
Budget & Random & World-space + SRD & No SRD & SRD gain [95\% CI]\\
\midrule
113,000 & 18.0134 & 21.0183 & 17.6617 & 3.3567 [3.1792, 3.5340]\\
226,000 & 19.0984 & 22.7899 & 21.9609 & 0.8290 [0.7672, 0.8925]\\
338,000 & 20.9305 & 23.1221 & 22.9514 & 0.1707 [0.1492, 0.1932]\\
Full & identity & 23.0695 & 22.9190 & 0.1505 [0.1163, 0.1847]\\
\bottomrule
\end{tabular}
\end{table}

\paragraph{Spatial responsibility decoding.}
For SRD, all branches start from a common dense-10K parent and train for a further 40,000 steps with the same seed, differing only in SRD enablement. SRD improves PSNR by 3.3567 dB at 113K, 0.8290 dB at 226K, and 0.1707 dB at 338K. It also improves every one of the 140 scenes at 113K and 226K. The paired 95\% bootstrap intervals remain positive across all reported budgets (Table~\ref{tab:ablation}), linking the largest gain to the most severe local compression and demonstrating the value of adapting Gaussian support after allocation.

\paragraph{Allocation before complete decoding.}
We compare decode-before-allocate (DBA) and budget-before-decoding (BBD) using identical selected candidates. At 113K, BBD reduces attribute projection from 1.308 to 0.411 ms and construction from 166.089 to 163.216 ms. Attribute-projection reductions are 68.5\%, 44.9\%, and 21.3\% at 113K, 226K, and 338K, respectively, because only selected candidates undergo complete attribute projection. The corresponding construction reductions are 1.73\%, 1.09\%, and 0.66\%. At Full, DBA and BBD require 96.75 and 96.65 ms because every candidate is retained. Appendix~\ref{app:eff_details} reports the full budget sweep.

\section*{Conclusion}
Mobile Gaussian reconstruction must construct an asset that a downstream device can store, transmit, render, and manipulate, rather than predict a dense primitive set at a fixed input-dependent size. We introduce \method, a feed-forward framework that constructs a Gaussian asset at an explicit representation budget. Its image-conditioned world-space latent field associates repeated multi-view observations in latent cells, whose local detail, view-mean evidence, and population determine exact integer capacity. A budget-conditioned decoder fuses cell information into retained latents, projects complete Gaussian attributes only for the selected support, and applies spatial responsibility decoding to adapt covariance under local sparsification. Results on DL3DV, Tanks \& Temples, MegaDepth, and on-device Mip-NeRF 360 demonstrate quality--asset-size operating points and direct mobile execution across user-selected budgets. These results establish world-space latent allocation as an effective formulation for feed-forward mobile Gaussian reconstruction.

\bibliographystyle{iclr2027_conference}
\bibliography{PocketSplat}

\section*{AI Use Statement}
Generative AI tools were primarily used to polish the manuscript's language and presentation. The authors are responsible for all technical content, experimental results, citations, and conclusions.

\appendix
\section{Mobile Export and Parity Requirements}
Export artifacts comprise preprocessing and post-allocation ONNX graphs, an FP32 pre-allocator Core ML package, and six DA3 Core ML stage packages. Export manifests should retain checkpoint/config hashes, exporter revision, precision, input shapes, model hashes, and interface names. Equivalent Core ML rewrites include reshape/expand in place of fixed meshgrids, elementwise products or matrix multiplication in place of some einsums, non-in-place pose decoding, and rank-5 cross-view aggregation. Their numerical behavior must be checked, especially near voxel boundaries.

\section{Report Consistency and Additional Experimental Details}
\label{app:reportaudit}
This appendix provides supplementary experimental details and implementation information.

\section{Allocation Implementation Details}
\label{app:allocation_impl}
Candidate detail is computed after restoring context RGB from ImageNet normalization, converting to luma $0.299R+0.587G+0.114B$, and quantizing to 256 levels. Within each cell, SHA256/splitmix64-derived pseudorandom keys use scene identity, a fixed seed, budget, and cell identity to select the quota. Reliability is normalized and clipped between its 20th and 95th percentiles before reliability--support weighted pooling.

\section{Hybrid iOS Deployment and Implementation Details}
\label{app:mobile_deployment}
\label{app:mobile_impl}
This section describes the on-device execution of geometry, allocation, fusion, decoding, and rendering.
\paragraph{Runtime partition.}
The PocketSplat deployment uses ONNX Runtime for preprocessing and post-allocation decoding, six Core ML stages for the backbone, Core ML FP32 for depth, cameras, and GS pre-allocation, Swift for voxel grouping, V2 allocation and hidden fusion, and Metal for viewing. Table~\ref{tab:abi} summarizes the tensor interfaces. For the cost-volume baselines, the common direct hybrid path exhausts memory. MVSplat becomes deployable after streaming its 32 depth candidates in chunks of eight and placing the Core ML backbone on CPU. DepthSplat remains out of memory after streamed cost-volume construction and staged execution.

\begin{table}[t]
\centering\small\setlength{\tabcolsep}{3.5pt}
\caption{Mobile deployment outcomes on the target iPhone under the four-view $256\times448$ FP32 protocol.}
\label{tab:mobile_deployment}
\begin{tabular}{llll}
\toprule
Method & Direct path & Final status & Mobile configuration\\
\midrule
PocketSplat & Deployable & Deployable & Core ML + ONNX Runtime\\
MVSplat & OOM & OOM & Direct cost-volume path\\
MVSplat-Stream & -- & Deployable & CPU Core ML + streamed cost volume ($D=8$)\\
DepthSplat & OOM & OOM & Streamed cost volume + staged execution\\
\bottomrule
\end{tabular}
\end{table}

\begin{table}[t]
\caption{Fixed mobile interfaces; batch size is one. $V=4$, $H=252$, $W=448$, $N=451{,}584$. All listed shapes include the batch dimension.}
\label{tab:abi}
\centering\small
\begin{tabular}{ll}
\toprule
Boundary / tensor & Shape \\
\midrule
Preprocess image / intrinsics / C2W & $[1,4,3,H,W]$ / $[1,4,3,3]$ / $[1,4,4,4]$ \\
Camera token & $[1,4,1024]$ \\
Backbone state / raw feature & $[1,4,577,1024]$ / $[1,4,577,2048]$ \\
Pre-allocator hidden / depth & $[1,4,32,H,W]$ / $[1,4,H,W]$ \\
Fused hidden to post graph & $[1,N,32]$ \\
Post means / covariance / opacity & $[1,N,3]$ / $[1,N,3,3]$ / $[1,N]$ \\
\bottomrule
\end{tabular}
\end{table}

\paragraph{Output budget versus decoding work.}
The post graph retains $N$ output rows at all budgets. Unselected entries receive opacity logit $-20$, and the PLY writer filters values $\le-19$. The selected count is exactly $B$ under the allocation constraints. Covariance is decomposed into scale and rotation for export. Current UI budgets are $451{,}584$, $338{,}000$, $225{,}792$, and $112{,}896$.

\paragraph{Camera contract.}
ARKit provides OpenGL camera-to-world poses. Conversion uses $C2W_{\rm cv}=C2W_{\rm gl}\operatorname{diag}(1,-1,-1,1)$. Images and intrinsics follow the GPU resize/crop transform. Preprocessing constructs cameras relative to the first view and normalizes translation using the median multi-view relative translation scale. Before export, similarity alignment maps predicted context cameras to the capture coordinate system. This contract preserves geometric consistency between captured multi-view images and the exported world-space Gaussian asset.

\section{Cross-Dataset Results and Protocol}
\label{app:ood_details}

Table~\ref{tab:main_all} reports frozen-checkpoint results on Tanks \& Temples and MegaDepth. Within \method, 226K gives the largest PSNR on both datasets (17.965 and 17.178 dB), Full gives the lowest LPIPS (0.3335 and 0.3450), and 338K gives the highest SSIM. \mbox{These results summarize cross-dataset quality across all reported budgets.}

At 338K, \method{} records 17.956 dB on Tanks \& Temples and 17.144 dB on MegaDepth. Table~\ref{tab:main_all} reports the complete cross-dataset results. \mbox{The intermediate budgets remain useful choices outside the training distribution.}

\section{Efficiency Protocols and Complete Measurements}
\label{app:eff_details}

\paragraph{Measurement protocol.}
We use the deployment efficiency round in the combined report on an NVIDIA GeForce RTX 4080 SUPER. Eight fixed scenes receive three warm-up and ten measured repeats per scene in a fresh process per arm. $T_{\mathrm{INFER50}}$ measures the GPU-resident interval from four input images and cameras to 50 rendered RGBs; loading, disk I/O, image decoding, CPU preprocessing, host-to-device copies, metric computation, and saving are excluded. We report the source medians. Peak memory is measured after load and warm-up in a separate process; values are reported in GB. The protocol fixes the timing interval and aggregation procedure for every evaluated GPU configuration. These fixed conditions isolate runtime variation from scene sampling and transient initialization effects. They also make the construction-time and peak-memory summaries reproducible under a single clearly defined profiling path.

\begin{table}[t]
\caption{GPU efficiency for the deployment path. Times are median $T_{\mathrm{INFER50}}$ in milliseconds and peak memory is reported in GB.}
\label{tab:efficiency}
\centering\small
\begin{tabular}{lrr}
\toprule
Method & $T_{\mathrm{INFER50}}$ (ms)$\downarrow$ & Peak (GB)$\downarrow$\\
\midrule
pixelSplat & 837.7 & 8.483\\
MVSplat & 322.8 & 13.885\\
DepthSplat & 306.2 & 7.519\\
TranSplat & 537.8 & 3.552\\
F$^4$Splat (113K) & 301.8 & 17.796\\
F$^4$Splat (226K) & 306.4 & 17.796\\
F$^4$Splat (338K) & 314.9 & 17.796\\
F$^4$Splat (Full) & 334.0 & 17.796\\
\method (113K) & 601.8 & 2.705\\
\method (226K) & 593.2 & 2.705\\
\method (338K) & 466.3 & 2.705\\
\method (Full) & 322.3 & 2.705\\
\bottomrule
\end{tabular}
\end{table}

\paragraph{Runtime and memory trade-offs.}
At Full, the deployment profiling path reports 322.3 ms and a 2.705 GB peak (Table~\ref{tab:efficiency}). Its total is close to MVSplat's 322.8 ms, while its reported peak is lower than the other methods in this table. \mbox{This profile records the complete GPU construction path at the dense operating point.}

The four \method{} rows in Table~\ref{tab:efficiency} show that reducing the output from Full to 113K increases $T_{\mathrm{INFER50}}$ from 322.3 to 601.8 ms while leaving peak memory at 2.705 GB. Dense features and allocation overhead dominate the savings from fewer rendered primitives. \mbox{The allocation-order ablation isolates the budget-dependent decoding effect.}

\paragraph{Construction effect of selecting before decoding.}
The paired experiment compares decode-before-allocate (DBA) and budget-before-decoding (BBD) with identical allocation and selected candidates, eight scenes, and 20 measured repeats. Table~\ref{tab:order} shows that skipping attribute projection on unselected rows reduces projection time by 68.5\% at 113K and construction time by 1.73\%. At that budget, allocation itself takes approximately 127 ms, versus 1.308 ms for dense projection.

\begin{table}[t]
\caption{Execution-order ablation. Values are median milliseconds per allocation call.}
\label{tab:order}
\centering\small\setlength{\tabcolsep}{4pt}
\begin{tabular}{rrrrrr}
\toprule
& \multicolumn{2}{c}{Attribute projection} & \multicolumn{2}{c}{Construction} & Construction\\
Budget & DBA & BBD & DBA & BBD & reduction\\
\midrule
113,000 & 1.308 & 0.411 & 166.089 & 163.216 & 1.730\%\\
226,000 & 1.289 & 0.710 & 174.178 & 172.276 & 1.092\%\\
338,000 & 1.308 & 1.029 & 180.567 & 179.378 & 0.659\%\\
Full & 1.29 & 1.32 & 96.75 & 96.65 & 0.096\%\\
\bottomrule
\end{tabular}
\end{table}

\section{Archived OOD Qualitative Panels}
\label{app:qualitative}
The following OOD panels are reproduced from the supplied experiment directory without altering individual renders. They preserve the original method labels. Their image-level provenance and budgets are not specified in the supplied summaries. They are included as archived illustrations, not controlled matched-budget evidence. The panels nevertheless provide a visual record of the supplied out-of-distribution reconstructions. They complement the quantitative transfer results with examples from the archived evaluation package. Their consistent method ordering offers an additional qualitative view of the representation budgets reported in the quantitative transfer evaluation.

\begin{figure}[htbp]
\centering
\includegraphics[width=\linewidth]{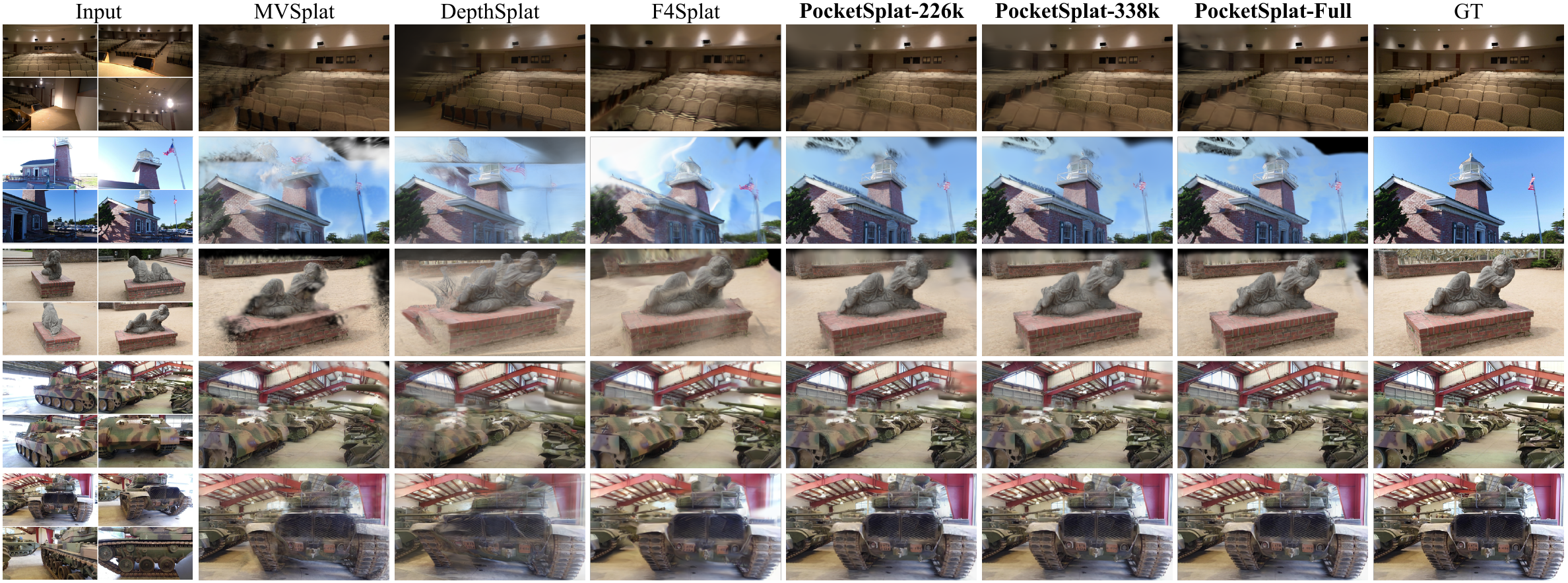}
\caption{MegaDepth qualitative comparison with five examples. Columns show context inputs, MVSplat, DepthSplat, F$^4$Splat, PocketSplat at 226K, 338K, and Full, and ground truth.}
\label{fig:qual_md}
\end{figure}

\end{document}